\documentclass{article}
\usepackage{spconf,amsmath,array,amssymb,graphicx}
\usepackage{lipsum}

\def\LL{{\cal L}}

\def\E{\mathbb{E}}

\def\M{\mathbf{M}}
\def\P{\mathbf{P}}

\usepackage{xcolor}

\usepackage{hyperref}

\title{CASCADE CONTEXT ENCODER FOR IMPROVED INPAINTING}

\name{Bartosz Zieli\'nski, \L{}ukasz Struski, Marek \'Smieja, Jacek Tabor\sthanks{The work from this paper were supported by the National Science Centre, Poland under grant agreement no 2015/19/B/ST6/01819.}}
\address{Faculty of Mathematics and Computer Science, Jagiellonian University, Krak\'ow, Poland}
\newtheorem{definition}{Definition}[section]

\begin{document}
%
\maketitle
\begin{abstract}
In this paper, we analyze if cascade usage of the context encoder with increasing input can improve the results of the inpainting. For this purpose, we train context encoder for 64x64 pixels images in a standard way and use its resized output to fill in the missing input region of the 128x128 context encoder, both in training and evaluation phase. As the result, the inpainting is visibly more plausible.

In order to thoroughly verify the results, we introduce normalized squared-distortion, a measure for quantitative inpainting evaluation, and we provide its mathematical explanation. This is the first attempt to formalize the inpainting measure, which is based on the properties of latent feature representation, instead of L2 reconstruction loss.
\end{abstract}
\begin{keywords}
inpainting, context encoder, quantitative inpainting evaluation, latent feature representation, normalized squared-distortion
\end{keywords}

\section{Introduction}
\label{sec:intro}

Image inpainting is the process of filling missing or corrupted regions in images based on surrounding image information so that the result looks visually plausible. It is widely used to rebuild damaged photographs, remove unwanted objects and texts, or replace objects.

Recently, deep learning techniques have been applied successfully to the problem of inpainting by Pathak et al. \cite{pathak2016context}. They introduced context encodersk CE (a convolutional neural network trained to generate the contents of an arbitrary image region conditioned on its surroundings), which are able to fill missing regions in natural images. Since its publication, various modifications of this method have been proposed \cite{zhao2017deep,yang2017high,li2017generative,zhang2018demeshnet}.

The overall architecture of context encoder is a simple encoder-decoder pipeline. The encoder takes an input image with missing regions and produces a latent feature representation of that image. The decoder takes this feature representation and produces the missing image content. According to results obtained in \cite{pathak2016context}, that kind of architecture works better for 64x64 pixels than for 128x128 pixels images, even though there is one additional layer in encoder and decoder in the case of the later. This is due to the fact that two additional layers are not enough to store the context information necessary to fill in four times greater missing region. It is worth mentioning, that this problem also appears in case of CE modifications.

\begin{figure}[t]
  \centering
  \includegraphics[width=8.5cm]{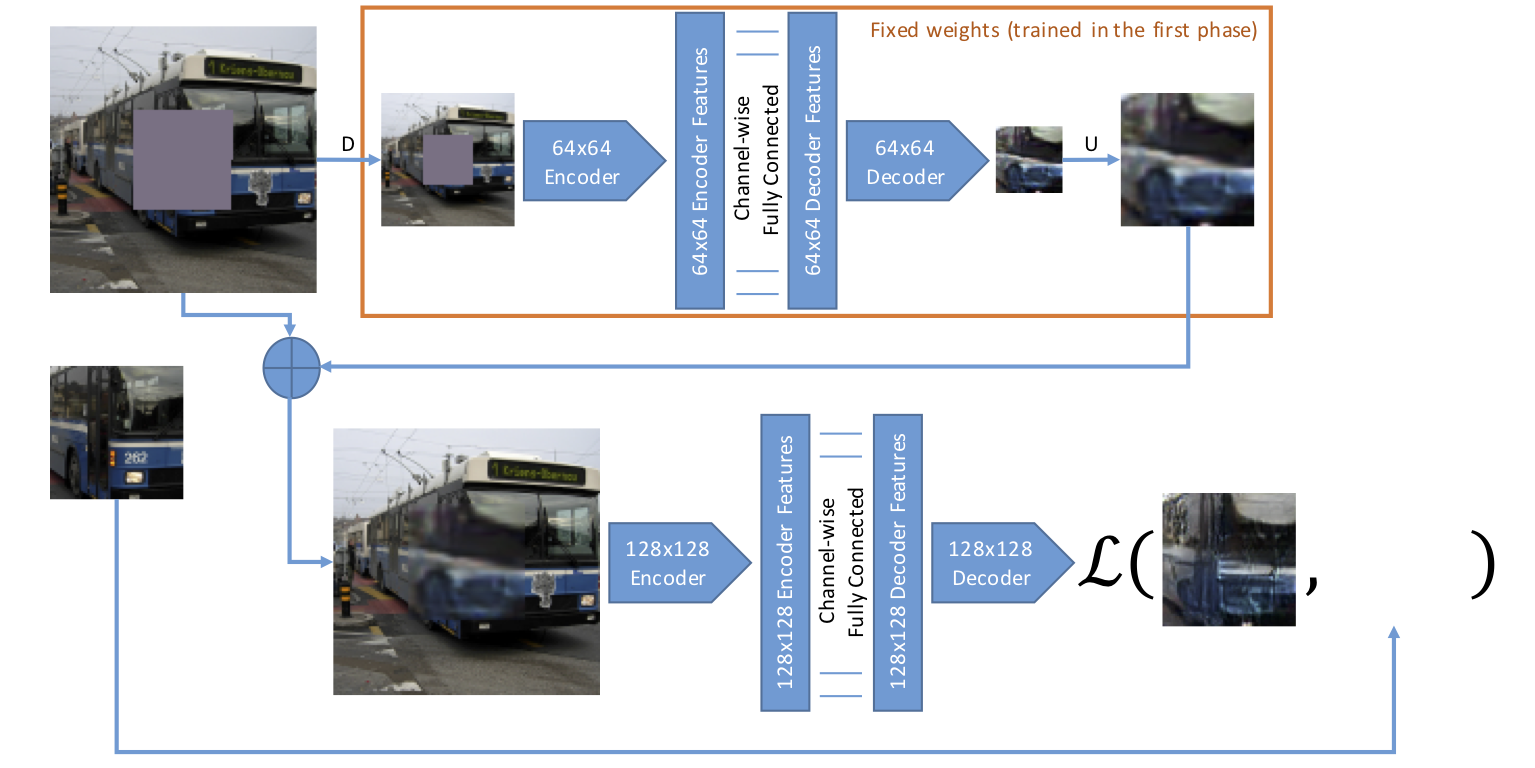}
\caption{Training cascade context encoder: The output of previously trained 64x64 context encoder is upscaled in order to fill in the missing (dropped out) input region of the 128x128 context encoder. As a result, the overall inpainting is produced by 64x64 CE, and 128x128 CE can concentrate on recovering the image details.}
\label{fig:scheme}
\end{figure}

In this paper, we analyze if cascade usage of the context encoder with increasing input can improve the results of the inpainting. For this purpose, we train 64x64 CE in a standard way and use its resized output in missing region to fill in the input of the 128x128 CE (see Fig.~\ref{fig:scheme}). In result, the overall inpainting is produced by 64x64 CE, and 128x128 CE can concentrate on recovering the image details.

The second, very important input of this paper is introducing theory of quantitative inpainting evaluation, based on latent feature representation. It is based on the intuition that particular image with different missing regions should be mapped to similar location in the latent space (see Fig. \ref{fig:methodology}). We call it \emph{normalized squared-distortion} (NSD) and we provide its mathematical explanation.

\section{Related works}
\label{sec:related}

Existing methods for inpainting problem can be divided into several categories such as structural inpainting~\cite{bertalmio2003simultaneous,liu2007image}, textures synthesis~\cite{bertalmio2000image,bertalmio2003simultaneous}, and example-based methods~\cite{criminisi2003object,criminisi2004region}. Structural inpainting uses geometric approaches to fill in the missing information in the region. Textures synthesis inpainting algorithms uses similar textures approaches, under the constraint that image texture should be consistent. Example-based image inpainting attempts to infer the missing region through retrieving similar patches or through learning-based model. The classical inpainting method can produce plausible output, however, they cannot handle hole-filling task, since the missing region is too large for local non-semantic methods to work well.

Over the recent years, convolutional neural networks have significantly advanced the image classification performance \cite{krizhevsky2012imagenet}. Motivated by the generative power of deep neural network, Pathak et al.~\cite{pathak2016context} used it as the backbone of their hole-filling approach. They introduced context encoder (CE), which is a type of conditional generative adversarial net, GAN~\cite{mirza2014conditional}. The overall architecture is a simple encoder-decoder pipeline, which is trained based on reconstruction and adversarial loss (the later obtained from discriminator).

This approach inspired many other researchers, and in result various modifications have been proposed. Yang et al.~\cite{yang2017high} proposed modification for high-resolution image inpainting, which uses two loss functions, the holistic content loss (conditioned on the output of the pre-trained content network) and the local texture loss (derived by pre-trained texture network). Li et al.~\cite{li2017generative} introduced additional local discriminator to distinguish the synthesize contents in the missing region (in contrast to global discriminator, which analyzes whole generated image). Moreover, they use the parsing network (pre-trained model which remains fixed) to ensure more photo-realistic images. Zhao et al.~\cite{zhao2017deep} proposed a cascade neural network, consisting two parts, where the result of inpainting GAN is further processed by deblurring-denoising network in order to remove the blur and noise.

In this paper, we propose an cascade architecture, which can be used not only with CE, but also with most of its modifications. We also introduce theory of quantitative inpainting evaluation, based on the properties of latent feature representation, which can be applied to any network with auto-encoder architecture.

\section{Theory of inpainting evaluation}
\label{sec:methodology}

Evaluation and comparison of the inpainting methods is very challenging task, because as we described in Section \ref{sec:intro} the main assumption is to obtain visually plausible results. The qualitative measures are highly subjective, while the quantitative measures (like PSNR) base only on L2 reconstruction loss. A a result, methods with blurred output are favored over potentially more plausible inpainting.

\begin{figure}[t]
  \centering
  \includegraphics[width=6cm]{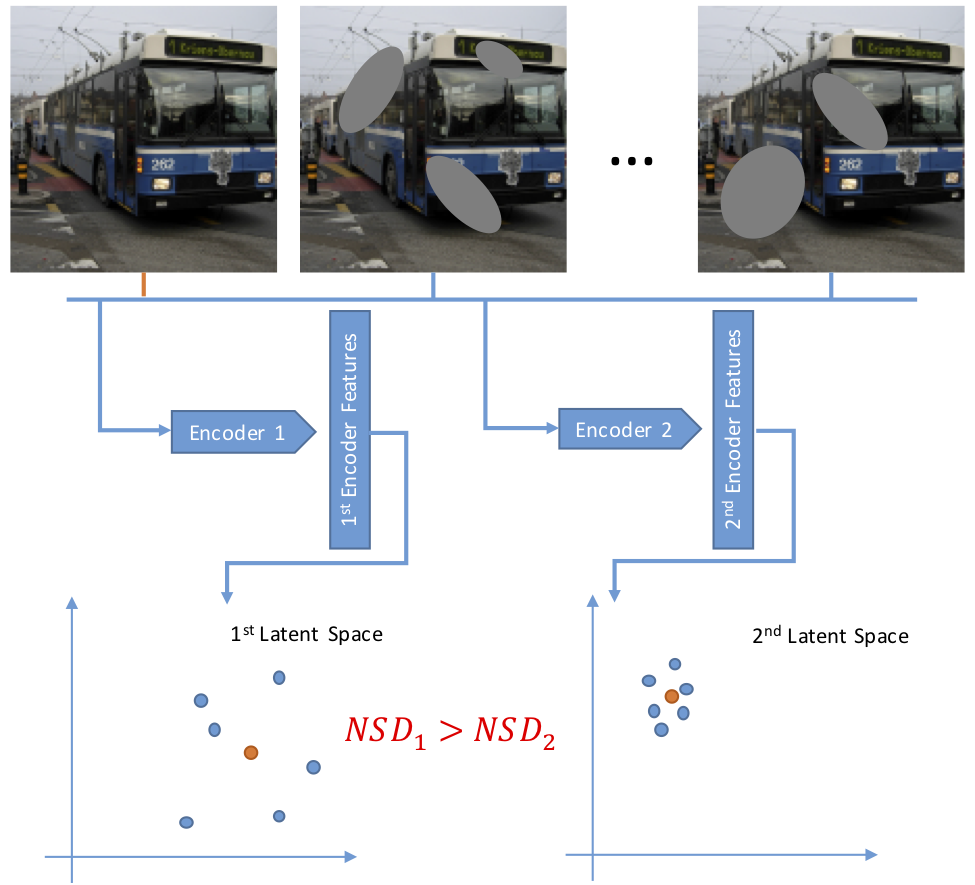}
\caption{In this figure one can observe the hypothetical latent feature vectors of two context encoders. The left context encoder has larger NSD than the right one, therefore it is less likely that results of inpainting will be similar for particular image with different missing regions.}
\label{fig:methodology}
\end{figure}

In this section, we formulate a theoretically based methodology for quantitative inpainting evaluation. This method bases on the assumption that latent feature vectors of the same image with different missing regions are closer to each other (right plot in Fig. \ref{fig:methodology}) in case of the context encoder with more plausible inpainting.

In the construction of this quantitative measure for inpainting evaluation we based on the following statements defined by \cite[Introduction]{pathak2016context}: (i) CE consists of an encoder capturing the context of an image into a compact latent feature representation and a decoder which uses that representation to produce the missing image content; (ii) CE needs to solve a much harder task than autoencoder, because it has to to fill in large missing areas of the image, where it can’t get ``hints'' from nearby pixels; (iii) CE requires a much deeper semantic understanding of the scene, and the ability to synthesize high-level features over large spatial extents. The main concept of CE was to use much larger dimension of the latent space ($D=4000$) than standard autoencoders ($D$ typically equals few hundreds) in order to obtain the latent features which are maximally resistant to the variability of the missing regions. Taking this into account, we argue that intuitively, a particular image with different masks should be mapped to similar location in the latent space.

Let $\P$ denote the random vector generating images, $\M$ random vector generating masks (various missing regions), and $E(P,M)$ denote our encoder which transports the image into $N(0,I)$. Thus we arrive at the following definition.

\begin{definition}
The mean squared-distortion of a given image $P$ generated from $\P$:
\[
\begin{split}
dist^2_E(P;\M) &= \E( \|E(P,M_1)-E(P,M_2)\|^2: M_1,M_2 \\
 & \mbox{ drawn independently from }\M).
\end{split}
\]
Now we define the mean square-distortion of $\P$:
$$
dist^2_E(\P,\M)=\E(dist_E^2(P,\M): P\mbox{ drawn from }\P).
$$
\end{definition}

If we have a sample of randomly drawn masks $(M_i)_{i=1..n}$ (in our experiment we take $n=100$) for a given $P$, we compute the estimator of $dist^2_E(P;\M)$:
$$
dist^2_E(P;(M_i)_i) = \frac{1}{n^2}\sum_{i,j=1..n}
\|E(P,M_i)-E(P,M_j)\|^2.
$$
To avoid the square complexity with respect to $n$, we apply the well-known equality
for arbitrary sequence $(x_i)_{i=1}^n$ \cite{spath1975cluster}:
$$
\frac{1}{2n^2} \sum_{i,j=1..n}\|x_i-x_j\|^2=
\frac{1}{n}\sum_{i=1..n} \|x_i-\bar x\|^2,
$$
where $\bar x$ denotes the mean of the sequence $(x_i)_i$.
Thus our final formula estimator is given by:
$$
dist^2_E(P;(M_i)_i):= 
\frac{2}{n}\sum_{i=1..n}
\|E(P,M_i)-\bar Z\|^2,
$$
where $\bar Z=\frac{1}{n}\sum_i E(P,M_i)$. 

Thus if we fix a sample $(P_j)_{j=1..k}$ (in our experiment we take $k=250$) of images, the estimator for 
$dist^2_E(\P,\M)$ is given by:
$$
dist^2_E((P_j)_j,(M_i)_i)=\frac{1}{k}
\sum_{j=1..k} dist_E^2(P_j,(M_i)_i).
$$

Our aim is to introduce the normalized measure, which would give $1$ if the change of the mask would make the representation in the latent space totally random, i.e. we consider the case when
$E(P,M_1)$ is totally independent of $E(P,M_2)$. In other words we want to consider how much we are better from the totally random distribution of $E(P,\M)$.

Let $X,Y$ be independent random vectors drawn from $N(0,I)$, where the aim is to compute
$$
E \|X-Y\|^2.
$$
Now $Z=X-Y$ is clearly a random vector such that $Z=(Z_1,\ldots,Z_D) \sim N(0,2I)$, and therefore $\frac{1}{\sqrt{2}}Z \sim N(0,I)$.

If $W_1,\ldots,W_D \sim N(0,I)$ are independent random variables, then $W_1^2+\ldots+W_D^2 \sim \chi^2_D$. Thus
$$
\frac{1}{2}\|Z\|^2 \sim \chi^2_D,
$$
and therefore $E\|X-Y\|^2=2D$.

Thus we finally arrive at the following definition.

\begin{definition}
The {\em normalized squared-distortion} is defined as 
$$
NSD(E;\P,\M)=\frac{1}{2D}dist^2_E(\P,\M).
$$
\end{definition}

Summarizing, given $k$-images $(P_j)_{j=1..k}$ and $n$-masks $(M_i)_{i=1..n}$, where the latent space is $D$-dimensional, the final 
estimator of the normalized square-distortion is given by
$$
\begin{array}{l}
NSD(E;(P_i)_i,(M_j)_j):=\\[1ex]
\displaystyle{\frac{1}{Dnk}\sum_{j=1..k}
\sum_{i=1..n}\|E(P_j,M_i)-\bar Z_j\|^2},
\end{array}
$$
where $\bar Z_j=\frac{1}{n}\sum_i E(P_j,M_i)$.

\begin{figure}[htb]
  \centering
  \includegraphics[width=7cm]{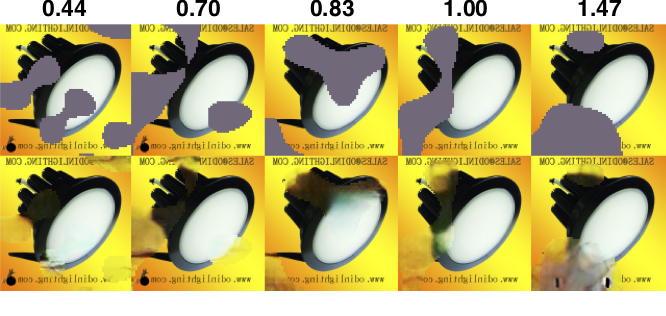}
  \includegraphics[width=7cm]{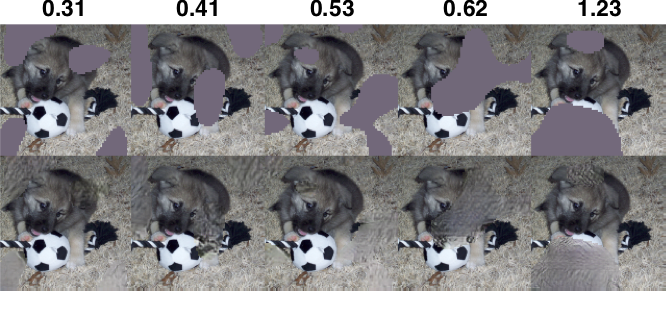}
\caption{The example images with different missing regions and values of normalized distance $\|E(P,M_i)-\bar Z\|^2/D$ (see supplemental materials at \url{http://www.ii.uj.edu.pl/~zielinsb} for more detailed samples).}
\label{fig:latent}
\end{figure}

The idea of this measure is presented in Fig.~\ref{fig:latent} for particular image $P$ and set of masks $(M_i)_i$.


\section{Cascade context encoder}
\label{sec:cascade}

In this section, we first give an overview of the context encoder, and then provide details on its cascade version. Moreover, we recall the learning procedure and various strategies for image region removal.

\subsection{Context encoder}

Context encoder is a simple encoder-decoder pipeline, taking image with missing regions as an input and produces the missing image content. However, in contrast to standard approaches, the encoder and the decoder are connected through a channel-wise fully-connected layer (in order to decrease the number of parameters). The encoder is derived from the AlexNet architecture \cite{krizhevsky2012imagenet}, trained for context prediction ``from scratch'' with randomly initialized weights. The channel-wise fully-connected layer is followed by a series of five up-convolutional layers \cite{dosovitskiy2017learning} of decoder.

Context encoders is trained by regressing to the ground truth content of the missing region. The L2 reconstruction loss is responsible for capturing the overall structure of the missing region, while adversarial loss tries to make prediction look real. The overall loss function is defined as:
\[
\LL = \lambda_{rec}\LL_{rec}+\lambda_{adv}\LL_{adv},
\]
where:
\[
\LL_{rec} = \left\lVert M \odot \Big( P - F \left( \overline{M} \odot P \right) \Big) \right\rVert_2
\]
and
\[
\begin{split}
\LL_{adv} &= max_D \E_{P \in \P} \bigg[log \Big(D(P) \Big) \\
         & + log \bigg(1 - D \Big(F \left(\overline{M} \odot P \right) \Big) \bigg) \bigg],
\end{split}
\]
where $P$ is an image, $F$ is an output of the context encoder, $M$ and $\overline{M}$ is binary mask corresponding to the missing regions and its complement, and $D$ is an output of the discriminator.

\subsection{Cascade context encoder}

As presented in Fig.~\ref{fig:scheme}, cascade context encoder contains two encoder-decoder pipelines, one for 64x64 CE and one for 128x128 CE. The first step is to train 64x64 CE using the same procedure like in \cite{pathak2016context}. In second step, 128x128 CE is trained and tested in the following manner: (i) the image region corresponding to mask $M$ is dropped; (ii) modified image is downscaled and forwarded with 64x64 CE; (iii) the filled region is upscaled and used to replace dropped region in original image; (iv) this image is forwarded with 128x128 CE. During the second step, the weights of 64x64 CE are fixed.

Let $F_{64}$ be the outputs of the 64x64 CE, and let $\searrow$ and $\nearrow$ be double downscaling and double upscaling operators. Then, assuming that 64x64 CE was already trained, the loss function for 128x128 CE is defined as:
\begin{equation}
\begin{split}
\LL_{rec} &= \| M \odot \bigg(P - F \Big( \overline{M} \odot P \\
         & + M \odot \nearrow{F_{64}} \left( \searrow{\overline{M}} \odot \searrow{P} \right) \Big) \bigg) \|_2
\end{split}
\label{eq:cceRec}
\end{equation}
and
\begin{equation}
\begin{split}
\LL_{adv} &= max_D \E_{P \in \P} \Bigg[ log(D(P)) \\
         & + log \Bigg( 1 - D \bigg( F \Big( \overline{M} \odot P \\
         & + M \odot \nearrow{F_{64}} \left( \searrow{\overline{M}} \odot \searrow{P} \right) \Big) \bigg) \Bigg) \Bigg].
\end{split}
\label{eq:cceAdv}
\end{equation}

\subsection{Region masks}

The input to a context encoder is an image with one or more of its regions ``dropped out'', the same like in \cite{pathak2016context}. The removed regions could be of any shape, we however test two strategies:
\begin{itemize}
\item Central region, where the region is the central square patch in the image, as shown in Fig. \ref{fig:scheme}.
\item Random blocks, where instead of choosing a single
large mask at a fixed location, we remove a number of smaller possibly overlapping masks, covering up to $1/4$ of the image (see Fig. \ref{fig:methodology}).
\end{itemize}

\section{Experiments}
\label{sec:experiments}

\begin{figure}[htb]
  \centering
  \centerline{\includegraphics[width=6.5cm]{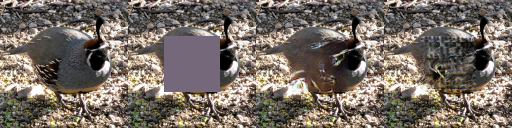}}
  \centerline{\includegraphics[width=6.5cm]{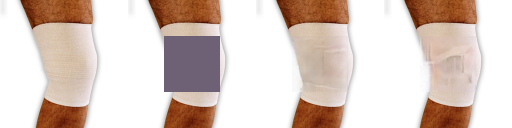}}
  \centerline{\includegraphics[width=6.5cm]{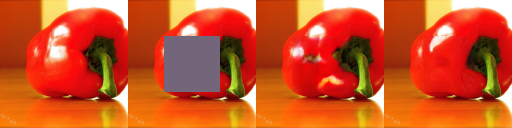}}
  \centerline{\includegraphics[width=6.5cm]{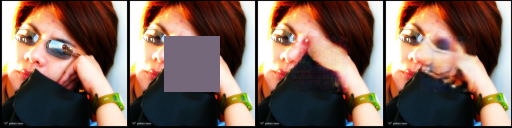}}
\caption{Visual comparison of results obtained for validation images when training with $100,000$ images from ImageNet. From left to right: original image, input image, CE, and our results. Two top images are the best results obtained for CE, while two bottom ones are the best inpaintings for CCE (see supplemental materials at \url{http://www.ii.uj.edu.pl/~zielinsb} for more results).}
\label{fig:res}
\end{figure}

We experimented with the subset of $100,000$ images from ImageNet dataset \cite{russakovsky2015imagenet} without using any of the accompanying labels. We trained context encoders with the joint loss function defined by Equation \ref{eq:cceRec} and \ref{eq:cceAdv} for the task of inpainting the missing region. We used the same encoder and discriminator architecture and the same parameters and hyper-parameters as in \cite{pathak2016context}.

Few qualitative results are shown in Fig.~\ref{fig:res} (more of them can be find in supplemental materials at \url{http://www.ii.uj.edu.pl/~zielinsb}). Our model performs generally better in inpainting semantic regions of an image (both central and random). One can observe that inpainting obtained with cascade approach has smoother borders. This is thanks to the fact that $64x64$ CE is responsible for returning overall inpainting, which usually fits better to the surrounding area. On the same time, $128x128$ CE can concentrate on image details, what results in more plausible inpainting.

Visually observed improvement was additionally confirmed by values of measure defined in Section \ref{sec:methodology}. $NSD$ for our approach equals $0.56 \pm 0.17$, while for standard CE it equals $0.79 \pm 0.17$ (for $k=250$ randomly chosen images and $n=100$ randomly chosen masks).

In the future research, we plan to include NSD training loss function. We believe that accurately modified loss function can produce network with smaller $NSD$, and in consequence improve the results of inpainting.

\bibliographystyle{IEEEbib}
\bibliography{missing}

\end{document}